\documentclass[conference]{IEEEtran}
\IEEEoverridecommandlockouts
\usepackage{cite}
\usepackage{amsmath,amssymb,amsfonts}
\usepackage{etoolbox,lipsum}
\usepackage{algorithmic}
\usepackage{graphicx}
\usepackage{textcomp}
\usepackage{xcolor}
\usepackage{booktabs} 
\usepackage{multirow}
\usepackage{graphicx}
\usepackage{caption}

\usepackage{multirow}
\usepackage{graphicx}
\usepackage{caption}
\usepackage{booktabs}
\usepackage{rotating}
\usepackage{array}
\usepackage{siunitx}
\usepackage{float}
\usepackage{overpic}
\usepackage{subcaption}
\usepackage{pgfplots}
\usepackage{siunitx}

\usepackage{rotating}
\usepackage{array}
\usepackage{siunitx}
\usepackage{float}
\usepackage{overpic}
\usepackage{subcaption}
\usepackage{cuted} 
\usepackage{pgfplots}
\def\BibTeX{{\rm B\kern-.05em{\sc i\kern-.025em b}\kern-.08em
    T\kern-.1667em\lower.7ex\hbox{E}\kern-.125emX}}
\pgfplotsset{compat=1.18} 
\begin{document}

\title{AugGS: Self-augmented Gaussians with Structural Masks for Sparse-view 3D Reconstruction}



\author{Bi'an Du, Lingbei Meng, Wei Hu* \thanks{* Corresponding author.} \\
Wangxuan Institute of Computer Technology, Peking University, Beijing, China \\
\texttt{\{pkudba,2100017808\}@stu.pku.edu.cn, forhuwei@pku.edu.cn} \\
}

\maketitle

\begin{strip}
  \centering
  \includegraphics[width=\textwidth]{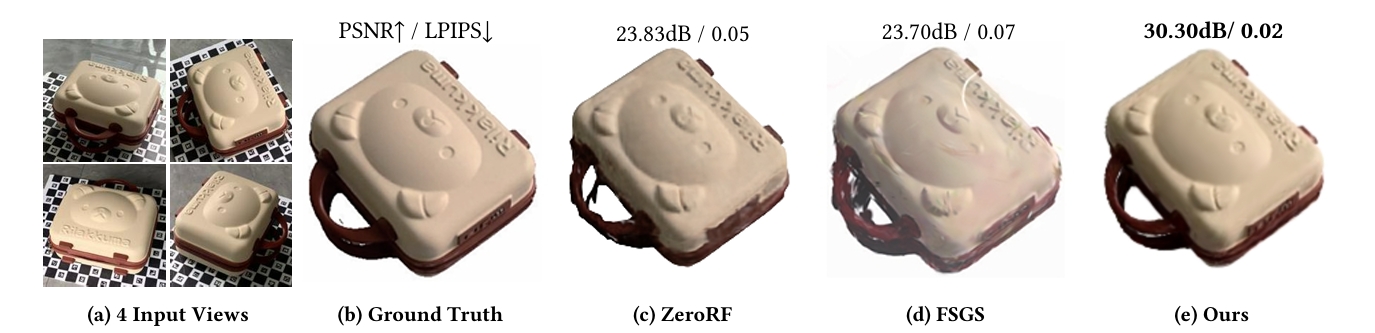}
  \captionof{figure}{Our method enables high-quality 3D reconstruction of sparse-view scenes with self-augmented Gaussian splatting, surpassing the current SOTA methods in both qualitative and quantitative aspects for sparse view 3D reconstructions.}
  \label{fig:teaser}
\end{strip}

\begin{abstract}

Sparse-view 3D reconstruction is a major challenge in computer vision, aiming to create complete three-dimensional models from limited viewing angles. Key obstacles include: 1) a small number of input images with inconsistent information; 2) dependence on input image quality; and 3) large model parameter sizes. To tackle these issues, we propose a self-augmented two-stage Gaussian splatting framework enhanced with structural masks for sparse-view 3D reconstruction. Initially, our method generates a basic 3D Gaussian representation from sparse inputs and renders multi-view images. We then fine-tune a pre-trained 2D diffusion model to enhance these images, using them as augmented data to further optimize the 3D Gaussians. Additionally, a structural masking strategy during training enhances the model’s robustness to sparse inputs and noise. Experiments on benchmarks like MipNeRF360, OmniObject3D, and OpenIllumination demonstrate that our approach achieves state-of-the-art performance in perceptual quality and multi-view consistency with sparse inputs.
\end{abstract}


\section{Introduction}
\label{sec:intro}

\begin{figure*}
  \centering
  \includegraphics[width=\textwidth]{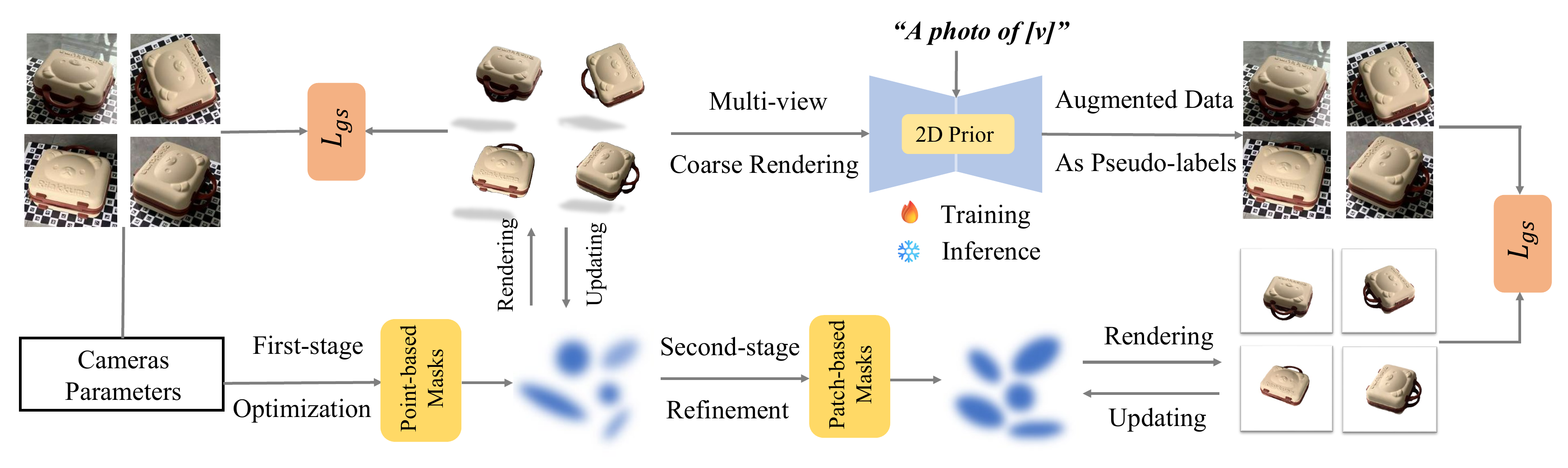}
  \caption{The overall architecture of our self-augmented Gaussian splatting method. We first create a coarse 3D Gaussian model from sparse-view images, generating a coarse point cloud and renderings from novel views. Multi-view renders and the 2D prior enhance perceptual quality, with structural masks integrated into the two-stage Gaussian process.}
  \label{fig:architecture}
\end{figure*}

    Reconstructing and rendering three-dimensional objects from two-dimensional images is a fundamental problem in the realm of computer vision. This intricate process involves translating sparse visual data into detailed and comprehensive 3D models, enabling applications such as augmented reality, robotics and 3D game/movie asset creation. Despite the potential, this task requires not only understanding complex geometries, but also typically capturing dozens of multi-view images, which is cumbersome and costly. Hence, it is desirable to efficiently reconstruct high-quality 3D objects from highly sparse captured images.

Previous methods primarily aim to reduce dependence on dense image captures by using auxiliary consistency loss to prevent degenerate reconstructions \cite{jain2021putting, Niemeyer2021Regnerf, shi2024zerorf, wang2023sparsenerf, zhu2023FSGS}. However, significant challenges remain with extremely sparse views, such as having only four images covering 360°. The limited input severely restricts detailed representation, often resulting in low-fidelity reconstructions and poor surface textures due to insufficient data consistency. Additionally, existing models heavily rely on the quality and number of input images, making them vulnerable to variations in object complexity and environmental noise, which highlights a robustness gap. Furthermore, the large parameter sizes of these models increase training burdens and reduce inference efficiency, thus limiting their practicality and scalability.

To address these challenges, we propose a novel two-stage Gaussian splatting method enhanced with perceptual data augmentation and structural masks for consistent 3D geometry and appearance reconstruction from sparse views.
In the first stage, a Gaussian model generates a basic 3D representation from limited inputs, serving as the initialization for the subsequent refinement. 
However, sparse views may lead to incomplete or blurry details in unobserved areas. To overcome this, we fine-tune a pre-trained image-conditioned ControlNet \cite{zhang2023adding} to enhance multi-view images. By leveraging the cross-pixel consistency of a pre-trained 2D diffusion model during denoising and detail restoration, we produce more reliable augmented data for the subsequent 3D Gaussian training, minimizing discrepancies between views. 
Additionally, we introduce structure-aware mask—point-based for the coarse Gaussian model and patch-based for the fine model to further improve robustness against sparse inputs and noise. This integrated approach streamlines the reconstruction process and enhances the quality and reliability of neural rendering in real-world scenarios.

Building on previous works \cite{yang2024gaussianobject, zhu2023FSGS, wang2023sparsenerf, shi2024zerorf}, we trained and evaluated our method on the MipNeRF360 \cite{Barron2021MipNeRF3U}, OmniObject3D \cite{wu2023omniobject3d}, and OpenIllumination \cite{liu2024openillumination} datasets. Our approach achieves state-of-the-art performance, outperforming the leading competitor \cite{yang2024gaussianobject} by 5–30\% in LPIPS, 8–25\% in PSNR, and 5–10\% in SSIM across various view counts and datasets. Additionally, our solution without 2D priors not only surpasses the top competitor in performance but also enhances the training efficiency by 20× and inference efficiency by 10\%.

Our contributions are summarized as follows: 
\begin{itemize}
    \item We introduce a two-stage Gaussian splatting method that generates pseudo-labels and utilizes 2D priors for perceptual data augmentation, enabling high-quality reconstruction of undersampled regions from sparse views.
    \item To address occlusions and unobserved areas, we implement point-based masks and patch-based masks for the two-stage Gaussians, respectively, which improve the robustness of the model against sparse input and noise.
    \item Experiments on benchmark datasets show that our method outperforms existing state-of-the-art methods in both qualitative and quantitative metrics. Additionally, we significantly improve training and inference efficiency without compromising the reconstruction quality.
\end{itemize}

\section{RELATED WORKS}
\subsection{Differentiable Point-based Rendering}
Recent advances in differentiable point-based rendering have revolutionized scene reconstruction. SynSin \cite{wiles2020synsin} enhanced point clouds with additional features and used CNNs to generate images, addressing issues like incompleteness and noise. Point-NeRF \cite{xu2023pointnerf} improved NeRF by rendering discrete neural points, reducing the computational cost of volumetric methods, though still requiring seconds for rendering. Point-Radiance \cite{zhang2023differentiable} introduced spherical harmonics and a novel splatting strategy for real-time rendering, offering better color interpolation and higher visual quality, especially in complex lighting.

Building on these advances, 3D Gaussian Splatting (3DGS) \cite{kerbl20233d} uses 3D Gaussian models for high-quality, rapid scene reconstructions. However, point-based methods still require dense point configurations, making sparse-view 360° reconstructions challenging. Our approach integrates object structure priors with point-based rendering, providing a robust solution for sparse-view reconstruction and setting a new benchmark for efficiency and quality in complex scenarios.

\subsection{Neural Rendering for Sparse View Reconstruction}

NeRF's original approach struggles with undersampled data in sparse-view reconstruction. Methods using SfM-derived visibility or depth \cite{deng2022depthsupervised} show promise but require closely aligned views and costly depth maps. Alternative approaches \cite{wang2023sparsenerf} use monocular depth estimation but often produce coarse results. Vision-language models \cite{pmlr-v139-radford21a} ensure semantic consistency but lack fine details. Methods like \cite{shi2024zerorf} capture overall appearance but miss details. 

The GaussianObject model \cite{yang2024gaussianobject} introduced structure-prior Gaussian initialization, enabling reconstruction from sparse inputs by reducing required views from over twenty (e.g., FSGS \cite{zhu2023FSGS}) to four. However, it and similar models struggle with sparse 360° views, often relying on SfM points, which limit performance with scarce data. Our approach enhances GaussianObject by integrating a robust framework for sparse setups, reducing dependence on precise point clouds and improving neural rendering’s real-world applicability.

\section{METHOD}

This section explains our self-augmented Gaussian splatting method, beginning with an overview of key ideas. We then outline the two-stage training process, perceptual data augmentation using 2D prior knowledge, and the integration of structural masks in the Gaussian process, followed by the training objective.

\subsection{Overview}
We start with sparse reference images $X^{\text{ref}}=\{x_i\}_{i=1}^N$ covering a $\ang{360}$ range, along with camera parameters $\Pi^{\text{ref}}=\{\pi_i\}_{i=1}^N$ and masks $M^{\text{ref}}=\{m_i\}_{i=1}^N$. The goal is a two-stage 3D Gaussian model $\mathcal{G}$ for photo-realistic rendering from any viewpoint. In the first stage, the model generates coarse point clouds and renderings, optimized by perceptual data augmentation as pseudo-labels. Point-based masks and patch-based masks are used for the two-stage Gaussians, respectively. The framework is shown in Fig. \ref{fig:architecture}.

\begin{figure*}[ht]
  \centering
  \resizebox{0.85\textwidth}{!}{
  \begin{tabular}{cccccccc}
   
    & \multicolumn{3}{c}{MipNeRF360} & \multicolumn{4}{c}{OmniObject3D} \\
    \rotatebox{90}{Gaussian Object} 
    & \includegraphics[width=0.13\linewidth]{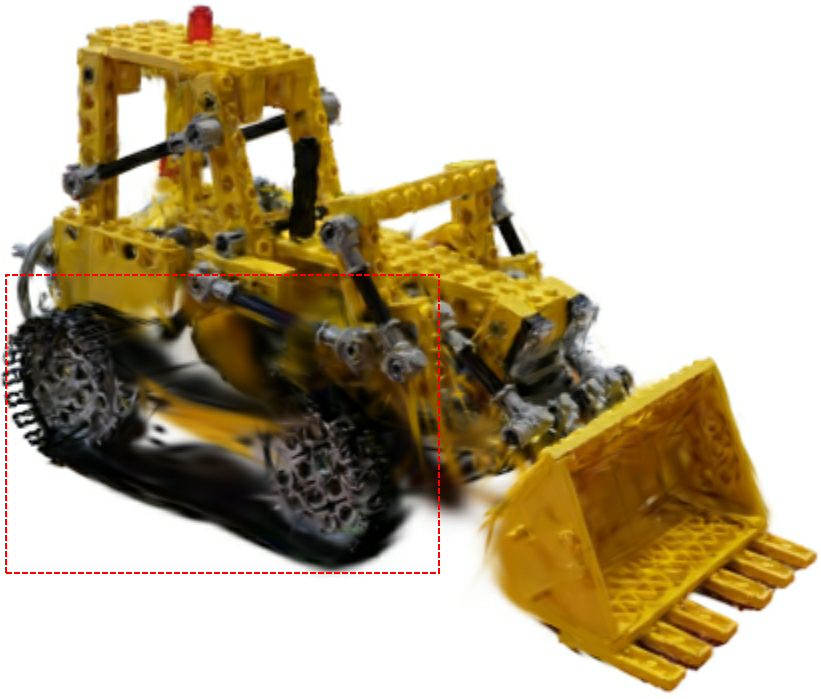}
    & \includegraphics[width=0.1\linewidth]{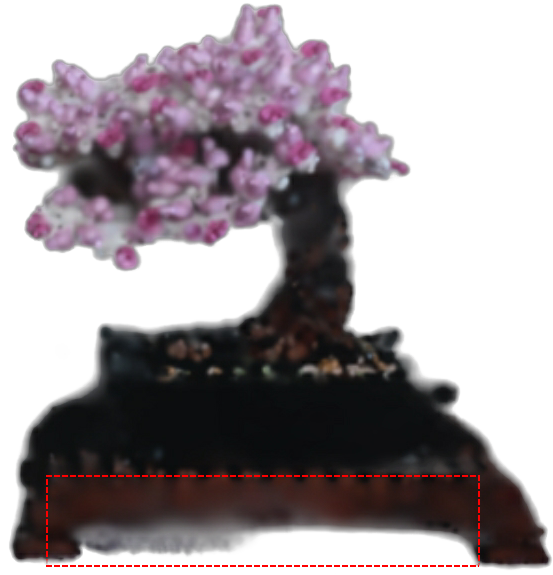}
    & \includegraphics[width=0.1\linewidth]{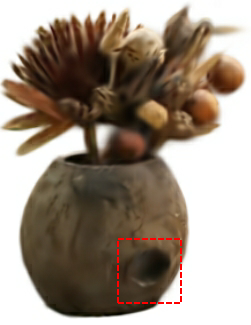}
    & \includegraphics[width=0.1\linewidth]{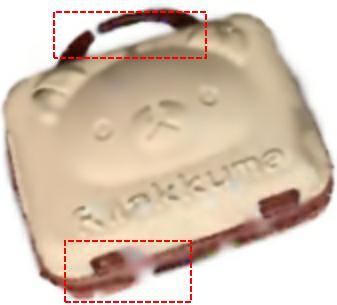}
    & \includegraphics[width=0.1\linewidth]{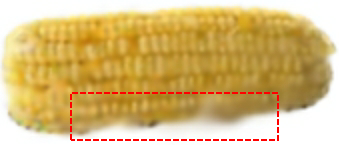}
    & \includegraphics[width=0.07\linewidth]{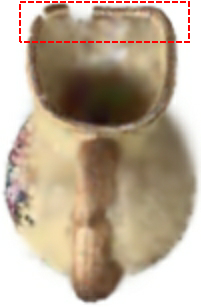}
    & \includegraphics[width=0.07\linewidth]{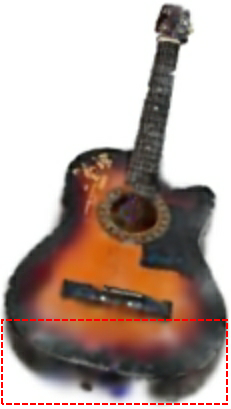} \\

    \rotatebox{90}{AugGaussian} 
    & \includegraphics[width=0.13\linewidth]{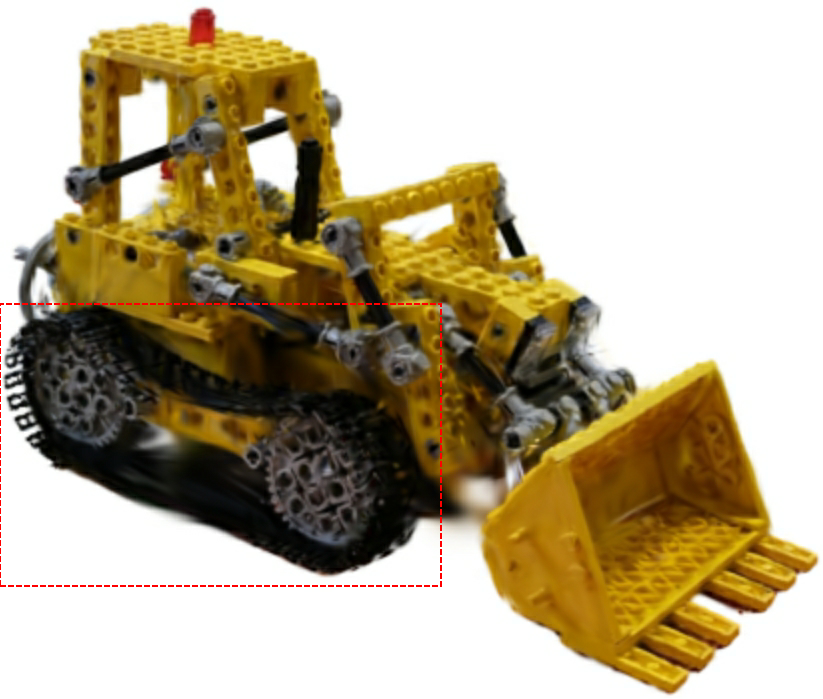}
    & \includegraphics[width=0.1\linewidth]{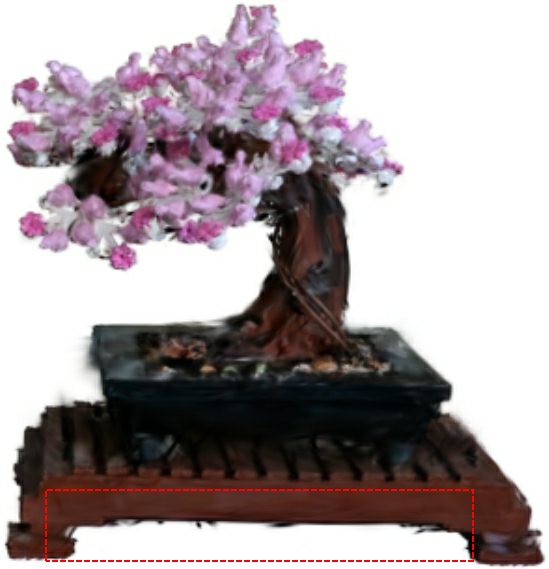}
    & \includegraphics[width=0.1\linewidth]{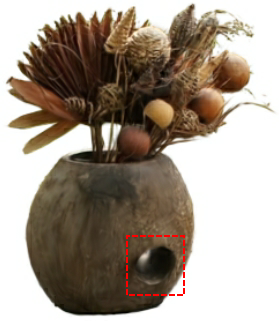}
    & \includegraphics[width=0.1\linewidth]{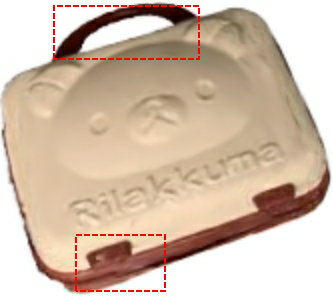}
    & \includegraphics[width=0.1\linewidth]{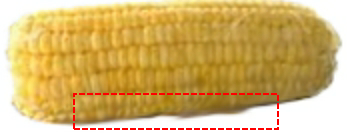}
    & \includegraphics[width=0.07\linewidth]{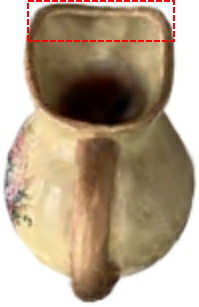}
    & \includegraphics[width=0.07\linewidth]{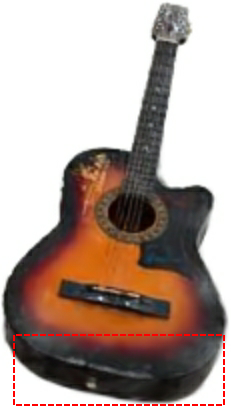} \\

    \raisebox{1.6\height}{\rotatebox{90}{GT}}
    & \includegraphics[width=0.13\linewidth]{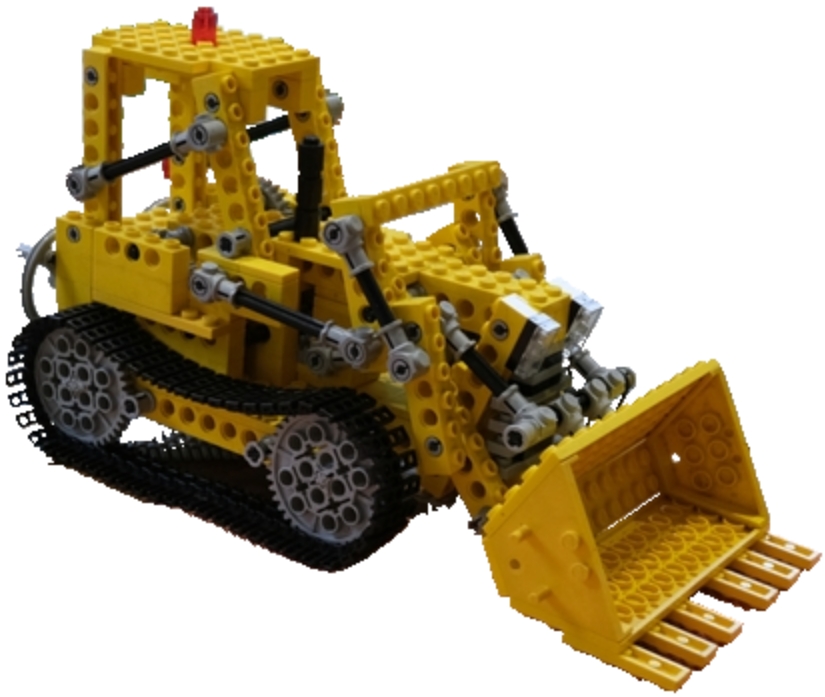}
    & \includegraphics[width=0.1\linewidth]{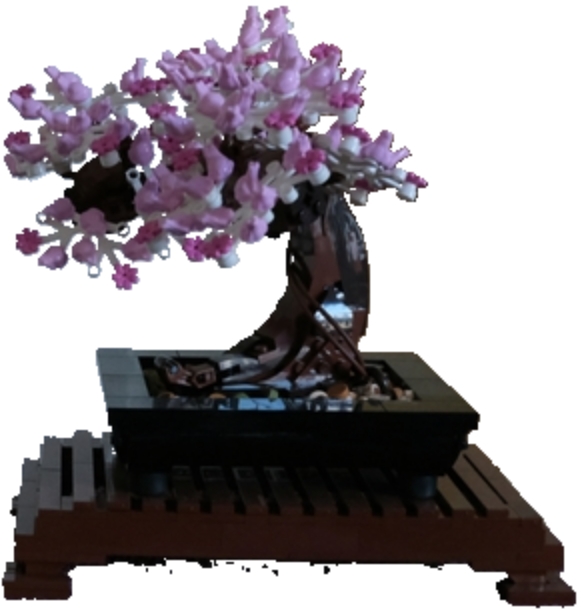}
    & \includegraphics[width=0.1\linewidth]{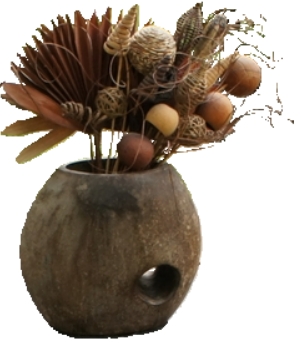}
    & \includegraphics[width=0.1\linewidth]{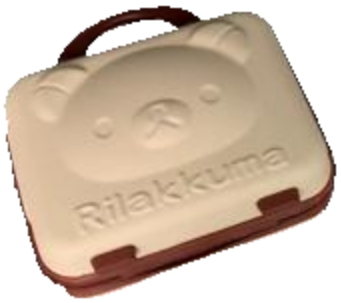}
    & \includegraphics[width=0.1\linewidth]{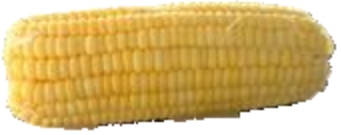}
    & \includegraphics[width=0.07\linewidth]{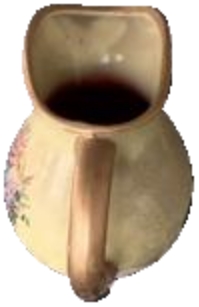}
    & \includegraphics[width=0.07\linewidth]{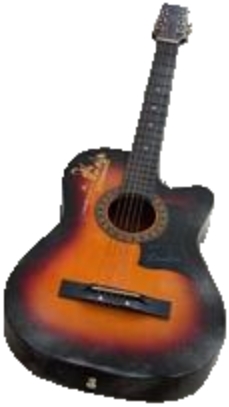} \\

  \end{tabular}
  }
  \caption{Qualitative examples on the MipNeRF360 and OmniObject3D dataset with 4 input views.}
  \label{tab:result image}
\end{figure*}

\begin{figure*}[h]
    \begin{tikzpicture}
    \begin{axis}[
        title={4View Data PSNR Comparison},
        xlabel={Iteration},
        ylabel={PSNR},
        xmin=0, xmax=11000,
        ymin=16, ymax=33,
        legend style={at={(0.5,-0.3)}, anchor=north, font=\footnotesize}, 
        legend columns=-1, 
        ymajorgrids=true,
        grid style=dashed,
        width=0.45\textwidth, 
        height=5cm, 
    ]    
    \addplot[color=red,mark=square] coordinates {
        (1000,25.918)     (2000,26.091)
    (3000,26.246)    (4000,26.279)
    (5000,26.361)    (6000,26.359)
    (7000,25.934)    (8000,25.443)
    (9000,24.969)    (10000,24.969)
    };
    \addlegendentry{K F}
    \addplot[    color=red,  mark=triangle, ] coordinates {
    (500,18.260)    (1000,23.204)    (1500,23.174)
    (2000,22.880)    (2500,22.819)
    (3000,23.068)    (3500,23.620)
    (4000,24.136)    (4500,23.980)
    (5000,24.109)    (5500,24.065)
    (6000,23.988)    (6500,23.988)
    (7000,23.988)    (7500,23.988)
    (8000,23.983)    (8500,23.983)
    (9000,23.947)    (9500,23.947)
    (10000,23.947)
    };
    \addlegendentry{K C}
    \addplot[color=cyan, mark=diamond,]
    coordinates {(1000,31.842)    (2000,31.787)
    (3000,31.749)    (4000,31.737)
    (5000,31.752)    (6000,31.794)
    (7000,31.534)    (8000,31.262)
    (9000,30.963)    (10000,30.963)};
\addlegendentry{G F}

\addplot[color=cyan, mark=*,]
    coordinates {(500,23.269)    (1000,25.926)    (1500,25.651)
    (2000,25.726)    (2500,25.735)
    (3000,25.713)    (3500,25.748)
    (4000,25.808)    (4500,25.895)
    (5000,25.833)    (5500,25.718)
    (6000,25.741)    (6500,25.741)
    (7000,25.793)    (7500,25.793)
    (8000,25.819)    (8500,25.819)
    (9000,25.831)    (9500,25.831)
    (10000,25.831)};
\addlegendentry{G C}

\addplot[color=teal, mark=x,]
    coordinates {(1000,23.281)    (2000,23.314)
    (3000,23.547)    (4000,23.592)
    (5000,23.626)    (6000,23.668)
    (7000,23.387)    (8000,23.047)
    (9000,22.722)    (10000,22.722)};
\addlegendentry{B F}

\addplot[color=teal, mark=+,]
    coordinates {(500,11.624)    (1000,18.456)
    (1500,19.243)    (2000,20.664)
    (2500,20.598)    (3000,20.745)
    (3500,21.234)    (4000,21.285)
    (4500,21.302)    (5000,21.528)
    (5500,21.519)    (6000,21.565)
    (6500,21.565)    (7000,21.587)
    (7500,21.587)    (8000,21.604)
    (8500,21.604)    (9000,21.610)
    (9500,21.610)    (10000,21.610)};
\addlegendentry{B C}

    \end{axis}
    \end{tikzpicture}
    \begin{tikzpicture}
    \begin{axis}[
        title={9View Data PSNR Comparison},
        xlabel={Iteration},
        ylabel={PSNR},
        xmin=0, xmax=10000,
        ymin=18, ymax=36,
        legend style={at={(0.5,-0.3)}, anchor=north, font=\footnotesize}, 
        legend columns=-1, 
        ymajorgrids=true,
        grid style=dashed,
        width=0.45\textwidth, 
        height=5cm, 
    ]
    
    \addplot[color=orange,mark=square] coordinates {
        (500,25.070)    (1000,27.695)
    (1500,28.047)    (2000,28.071)
    (2500,28.143)    (3000,28.193)
    (3500,28.394)    (4000,28.480)
    (4500,28.486)    (5000,28.368)    (5500,28.416)
    (6000,28.574)    (6500,28.223)
    (7000,27.882)    (7500,27.570)
    (8000,27.259)    (8500,27.259)
    (9000,27.259)    };
    \addlegendentry{K F}
    \addplot[color=orange,mark=triangle,]
    coordinates {
    (500,19.397)    (1000,27.822)
    (1500,27.849)    (2000,27.977)
    (2500,27.978)    (3000,27.644)
    (3500,27.606)    (4000,27.415)
    (4500,27.481)    (5000,27.206)
    (5500,27.318)    (6000,27.147)
    (6500,27.147)    (7000,27.044)
    (7500,27.044)    (8000,26.913)
    (8500,26.913)    (9000,26.750)
    (9500,26.750)    
    };
    \addlegendentry{K C}

    \addplot[color=blue, mark=diamond,]
    coordinates {(500,30.334)    (1000,33.342)
    (1500,33.504)    (2000,33.481)
    (2500,33.482)    (3000,33.608)
    (3500,33.593)    (4000,33.632)
    (4500,33.668)    (5000,33.713)
    (5500,33.636)    (6000,33.649)
    (6500,33.472)    (7000,33.310)
    (7500,33.099)    (8000,32.912)
    (8500,32.912)    (9000,32.912)};
\addlegendentry{G F}

\addplot[color=blue, mark=*,]
    coordinates {(500,19.387)    (1000,29.555)
    (1500,29.390)    (2000,29.462)
    (2500,29.554)    (3000,29.570)
    (3500,30.416)    (4000,31.098)
    (4500,32.559)    (5000,32.444)
    (5500,33.157)    (6000,32.986)
    (6500,32.986)    (7000,32.951)
    (7500,32.951)    (8000,32.877)
    (8500,32.877)    (9000,32.767)
    (9500,32.767)    };
\addlegendentry{G C}
\addplot[color=green, mark=x,]
    coordinates {(500, 21.763)    (1000, 25.679)
    (1500, 25.962)    (2000, 25.756)
    (2500, 25.825)    (3000, 25.980)
    (3500, 26.132)    (4000, 26.102)
    (4500, 26.068)    (5000, 26.024)
    (5500, 26.169)    (6000, 26.158)
    (6500, 25.904)    (7000, 25.569)
    (7500, 25.327)    (8000, 25.075)
    (8500, 25.075)    (9000, 25.075)};
\addlegendentry{B F}

\addplot[color=green, mark=+,]
    coordinates {(500, 13.502)    (1000, 19.697)
    (1500, 20.068)    (2000, 20.174)
    (2500, 20.562)    (3000, 21.565)
    (3500, 21.405)    (4000, 22.301)
    (4500, 22.787)    (5000, 23.705)
    (5500, 24.220)    (6000, 24.563)
    (6500, 24.563)    (7000, 24.571)
    (7500, 24.571)    (8000, 24.573)
    (8500, 24.573)    (9000, 24.545)
    (9500, 24.545)    };
\addlegendentry{B C}    
    \end{axis}
    \end{tikzpicture}
    
    \caption{Comparative analysis of PSNR metrics for 4View and 9View configurations across different objects and Gaussian iteration processes. Note: 'K', 'G', and 'B' represent objects Kitchen, Garden, and Bonsai, respectively. 'C' refers to the Coarse Gaussian iteration process, while 'F' denotes the Fine Gaussian iteration process.}
    \label{fig:iter}
\end{figure*}

\begin{figure}[]
 
    \begin{minipage}[c]{0.36\linewidth}
    \vspace{0.2cm}
    \fbox{\includegraphics[width=1\linewidth]{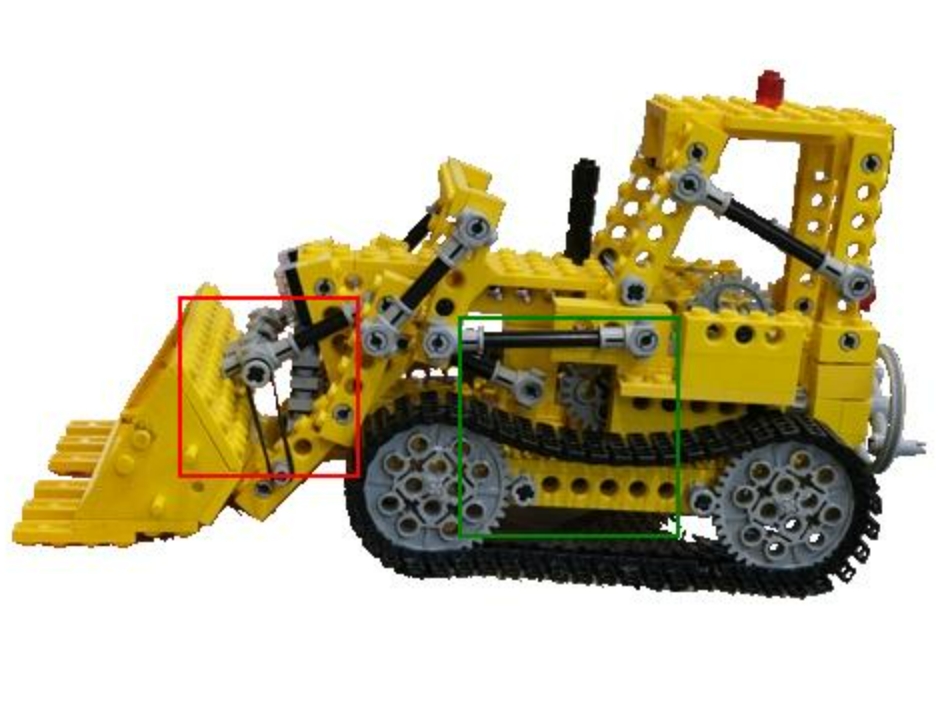}}
\end{minipage}
\hfill
\begin{minipage}[t]{0.58\linewidth}
    \begin{subfigure}[t]{0.25\linewidth}
    \setlength{\fboxsep}{0pt}
        \fcolorbox{green}{transparent}{\includegraphics[width=\linewidth]{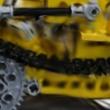}}
    \end{subfigure}%
    \hfill
    \begin{subfigure}[t]{0.25\linewidth}
    \setlength{\fboxsep}{0pt}
    \fcolorbox{green}{transparent}
        {\includegraphics[width=\linewidth]{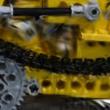}}
    \end{subfigure}%
    \hfill
    \begin{subfigure}[t]{0.25\linewidth}
    \setlength{\fboxsep}{0pt}
    \fcolorbox{green}{transparent}{
        \includegraphics[width=\linewidth]{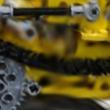}}
    \end{subfigure}%
    \hfill
    \begin{subfigure}[t]{0.25\linewidth}
    \setlength{\fboxsep}{0pt}
    \fcolorbox{green}{transparent}{
        \includegraphics[width=\linewidth]{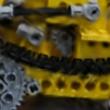}}
    \end{subfigure}
    
    \begin{subfigure}[t]{0.25\linewidth}
    \setlength{\fboxsep}{0pt}
    \fcolorbox{red}{transparent}{
        \includegraphics[width=\linewidth]{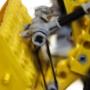}}
        \caption*{w/o Aug}
    \end{subfigure}%
    \hfill
    \begin{subfigure}[t]{0.25\linewidth}
    \setlength{\fboxsep}{0pt}
    \fcolorbox{red}{transparent}{
        \includegraphics[width=\linewidth]{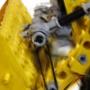}}
        \caption*{w/o PV}
    \end{subfigure}%
    \hfill
    \begin{subfigure}[t]{0.25\linewidth}
    \setlength{\fboxsep}{0pt}
    \fcolorbox{red}{transparent}{
        \includegraphics[width=\linewidth]{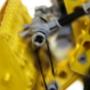}}
        \caption*{w/o M}
    \end{subfigure}%
    \hfill
    \begin{subfigure}[t]{0.25\linewidth}
    \setlength{\fboxsep}{0pt}
    \fcolorbox{red}{transparent}{
        \includegraphics[width=\linewidth]{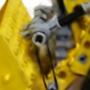}}
        \caption*{Ours}
    \end{subfigure}
\end{minipage}

  \caption{ Ablation study on different augmentation strategies. “Aug” denotes for augmentation, “PV” denotes for perceptual view augmentation and "M" for mask augmentation.}
  \label{fig:mask}
\end{figure}

\subsection{Two-stage Training Gaussians}

Sparse views, especially with only four images, offer minimal 3D information. In such cases, essential SFM points for initializing 3DGS are often lost, leading to missing geometry and reduced accuracy and efficiency in 3D Gaussian densification. 

To address geometric defects and density gaps in sparse views, the Gaussian point distribution is refined stage by stage. Using perceptual-level data augmentation, we guide point distribution in sparse regions, improving geometric consistency and density. In the first stage, we train a set of coarse Gaussian points, focusing on overall geometry and providing a spatial framework for maintaining structural coherence in unobserved regions. The coarse point cloud $\mathcal{P}_\text{c}=\{p_i: \mu_i, q_i, s_i, \sigma_i, sh_i \}_{i=1}^{P}$ consists of Gaussian points with parameters: center $\mu$, rotation quaternion $q$, scaling vector $s$, opacity $\sigma$, and SH coefficient $sh$.

In the second stage, Gaussian attributes are refined with perceptual data augmentation, improving geometric consistency and filling sparse regions. We convert spherical harmonics to color, keeping only spatial coordinates $\mu_i$ and color $c_i$, while excluding other attributes. This focus enhances geometry without adding complexity, simplifying model precision.

\subsection{Perceptual View data Augmentation}

3D Gaussians from sparse views often show blurriness or incoherence due to incomplete perspective information. However, a pre-trained 2D prior can restore missing details, creating more realistic images. We propose using multi-view coarse renders and the 2D prior to enhance perceptual views.

Following \cite{yang2024gaussianobject}, we add 3D noise $\epsilon_s$ to Gaussian attributes, as the leave-one-out approach is slow with little improvement. Degraded images are rendered from noisy Gaussians, and LoRA weights fine-tune a pre-trained Control-Net \cite{zhang2023adding} using these images, based on Stable Diffusion v1.5 \cite{rombach2022high}.

We render images $X^{\text{c}}=\{x_j\}_{j=1}^{N^{\prime}}$ from the first-stage Gaussians in $N^{\prime}$ novel views. These coarse images are input into the fine-tuned 2D prior model to generate restored images $\Tilde{X}^{\text{c}}$, and we augment the reference images to $X^{\text{aug}}=X^{\text{ref}}\cup \Tilde{X}^{\text{c}}$ with corresponding camera parameters $\Pi^{\text{aug}}=\Pi^{\text{ref}}\cup \Pi^{\text{c}}$. A second-stage training on the Gaussians yields photo-realistic renderings from any viewpoint, $x=\mathcal{G}_f(\pi|\{x_i, \pi_i\}_{i=1}^{N+N^{\prime}})$. Our approach, relying solely on the provided dataset without external visual data, ensures high accuracy and consistency in the 3D model.

The fine-tuned 2D prior model optimizes multi-view images from the second-stage Gaussians, improving perceptual quality. Even without post-processing, our method outperforms \cite{yang2024gaussianobject}, accelerating training and inference. With post-processing, it matches their speeds while further enhancing reconstruction quality (Tab.~\ref{tab:comparisons}).

The optimization of the first-stage Gaussians $\mathcal{G}_c$ and the second-stage Gaussians $\mathcal{G}_f$ incorpoates color and monocular depth losses. The color loss combines L1 and D-SSIM lossses:
\begin{equation}
    \mathcal{L}_\text{1}=||x-x^{\text{ref}}||_{1}, \mathcal{L}_{\text{D-SSIM}}=\text{1-SSIM}(x,x^{\text{ref}}),
\end{equation}
where $x$ is the rendering and $x^{\text{ref}}$ is the corresponding reference image.
A shift and scale invariant depth loss is utilized to guide geometry:
\begin{equation}
    \mathcal{L}_\text{d}=||D^*-D_{\text{pred}}^*||_{1},    
\end{equation}
where $D^*$ and $D^*_{\text{pred}}$ are per-frame rendered depths and monocularly estimated depths \cite{bhat2023zoedepth} respectively. The depth value are computed following a normalization strategy \cite{ranftl2020robust}:
\begin{equation}
    D^*=\frac{D-\text{median}(D)}{\frac{1}{M}\Sigma_{i=1}^M|D-\text{median}(D)|},
\end{equation}
where $M$ denotes the number of valid pixels. The overall loss combines these components as
\begin{equation}
    \mathcal{L}_{\text{gs}}=(1-\lambda_{\text{SSIM}})\mathcal{L}_\text{1} +\lambda_{\text{SSIM}}\mathcal{L}_{\text{D-SSIM}}+\lambda_{\text{d}}\mathcal{L}_\text{d},
\end{equation}
where $\lambda_{\text{SSIM}}$ and $\lambda{\text{d}}$ control the magnitude of each term.

\subsection{Integration of Structure-aware Masks in Coarse-to-Fine Gaussian Process}


\subsubsection{Point-based Masks For the First Stage}

We apply point-based random masks to 3D Gaussians to enhance the model's robustness to sparse inputs and noise. Specifically, we set a mask ratio $r_c$ and an update interval $t_c$. Every $t_c$ iterations, we randomly select $r_c*P$ 3D points to discard, preventing them from participating in the densification process.

In the first-stage Gaussian model, point-based masks improve structural accuracy by targeting Gaussian centers from sparse data. These masks are dynamically generated by analyzing rendering errors against real images, identifying regions with significant divergence. They direct focus to areas with higher errors, guiding adjustments to Gaussian parameters and conserving resources on well-represented regions, setting the stage for detailed refinement.

\subsubsection{Patch-based Masks For the Second Stage}

Sparse-view images often lack information from certain object parts, like a side of the kitchen. In the SFM point cloud, this appears as missing or sparse points, which is challenging to fully address during densification.

To enhance the second-stage 3D Gaussian model's robustness to missing data and noise, we use patch-based random masks. We extract $C$ patch centers via farthest point sampling (FPS) and generate $C$ patches using the kNN algorithm. With an iteration gap $t_f$ and mask ratio $r_f$, every $t_f$ iterations, $r_f*C$ patches are randomly selected and excluded from the densification process.
These masks refine second-stage Gaussian parameters, enhancing resolution in key areas. This targeted approach improves perceptual quality, using iterative feedback for high-fidelity reconstruction while preserving authenticity without external data.

\section{EXPERIMENT}

\begin{table*}
\caption{Comparisons on MipNeRF360 and OmniObject3D datasets with varying input views. LPIPS* = LPIPS x \(10^2\) throughout this paper. PP represents the post-processing step using the fine-tuned 2D prior model to optimize the final output images.}
\label{tab:comparisons}
\centering
\resizebox{0.8\textwidth}{!}{
\begin{tabular}{@{}c@{}c@{}cccccccccc@{}}
\toprule
& & \multicolumn{1}{c}{Method} & \multicolumn{3}{c}{4-view} & \multicolumn{3}{c}{6-view} & \multicolumn{3}{c}{9-view} \\ 
\cmidrule(r){4-6} \cmidrule(lr){7-9} \cmidrule(l){10-12}
& & & LPIPS* ↓ & PSNR ↑ & SSIM ↑ & LPIPS* ↓ & PSNR ↑ & SSIM ↑ & LPIPS* ↓ & PSNR ↑ & SSIM ↑ \\
\midrule
\multirow{10}{*}{\rotatebox[origin=c]{90}{MipNeRF360}} & & DVGO \cite{sun2022direct} & 24.43 & 14.39 & 0.7912 & 26.67 & 14.30 & 0.7676 & 25.66 & 14.74 & 0.7842 \\
& & 3DGS \cite{kerbl20233d} & 10.80 & 20.31 & 0.8991 & 8.38 & 22.12 & 0.9134 & 6.42 & 24.29 & 0.9331 \\
& & DietNeRF \cite{jain2021putting} & 11.17 & 18.90 & 0.8971 & 6.96 & 22.03 & 0.9286 & 5.85 & 23.55 & 0.9424 \\
& & RegNeRF \cite{Niemeyer2021Regnerf} & 20.44 & 13.59 & 0.8476 & 20.72 & 13.41 & 0.8418 & 19.70 & 13.68 & 0.8517 \\
& & FreeNeRF \cite{10205351} & 16.83 & 13.71 & 0.8534 & 6.84 & 22.26 & 0.9332 & 5.51 & 27.66 & 0.9485 \\
& & SparseNeRF \cite{wang2023sparsenerf} & 17.76 & 12.83 & 0.8454 & 19.74 & 13.42 & 0.8316 & 21.56 & 14.36 & 0.8235 \\
& & ZeroRF \cite{shi2024zerorf} & 19.88 & 14.17 & 0.8188 & 8.31 & 24.14 & 0.9211 & 5.34 & 27.78 & 0.9460 \\
& & FSGS \cite{zhu2023FSGS} & 9.51 & 21.07 & 0.9097 & 7.69 & 22.68 & 0.9264 & 6.06 & 25.31 & 0.9397 \\
& & GaussianObject \cite{yang2024gaussianobject} & 4.98 & 24.81 & 0.9350 & 3.63 & 27.00 & 0.9512 & 2.75 & 28.62 & 0.9638 \\
\cmidrule{2-12}
& & Ours (w/o PP) & 4.05 & 27.85 & 0.9515 & 3.27 & 29.12 & 0.9621 & 2.67 & 30.25 & 0.9701 \\
& & Ours (full setting)  & \textbf{3.85} & \textbf{28.46} & \textbf{0.9587} & \textbf{3.14} & \textbf{30.02} & \textbf{0.9685} & \textbf{2.60} & \textbf{30.96} & \textbf{0.9732} \\
\cmidrule{2-12}
\multirow{10}{*}{\rotatebox[origin=c]{90}{OmniObject3D}} & & DVGO \cite{sun2022direct} & 14.48 & 17.14 & 0.8952 & 12.89 & 18.32 & 0.9142 & 11.49 & 19.26 & 0.9302 \\
& & 3DGS \cite{kerbl20233d} & 8.60 & 17.29 & 0.9299 & 7.74 & 18.29 & 0.9378 & 6.50 & 20.26 & 0.9483 \\
& & DietNeRF \cite{jain2021putting} & 11.64 & 18.56 & 0.9205 & 10.39 & 19.07 & 0.9267 & 10.32 & 19.26 & 0.9258 \\
& & RegNeRF \cite{Niemeyer2021Regnerf} & 16.75 & 15.20 & 0.9091 & 14.38 & 15.80 & 0.9207 & 10.17 & 17.93 & 0.9420 \\
& & FreeNeRF \cite{10205351} & 8.28 & 17.78 & 0.9402 & 7.32 & 19.02 & 0.9464 & 7.25 & 20.35 & 0.9467 \\
& & SparseNeRF \cite{wang2023sparsenerf} & 17.47 & 15.22 & 0.8921 & 21.71 & 15.86 & 0.8935 & 23.76 & 17.16 & 0.8947 \\
& & ZeroRF \cite{shi2024zerorf} & 4.44 & 27.78 & 0.9615 & 3.11 & 31.94 & 0.9731 & 3.10 & 32.93 & 0.9747 \\
& & FSGS \cite{zhu2023FSGS} & 6.25 & 24.71 & 0.9545 & 6.05 & 26.36 & 0.9582 & 4.17 & 29.16 & 0.9695 \\
& & GaussianObject \cite{yang2024gaussianobject} & 2.07 & 30.89 & 0.9756 & 1.55 & 33.31 & 0.9821 & 1.20 & 35.49 & 0.9870 \\
\cmidrule{2-12}
& & Ours (w/o PP) & 1.87 & 32.59 & 0.9796 & 1.54 & 33.75 & 0.9844 & 1.18 & 35.51 & 0.9876 \\
& & Ours (full setting)  & \textbf{1.75} & \textbf{34.02} & \textbf{0.9834} & \textbf{1.48} & \textbf{34.19} & \textbf{0.9912} & \textbf{1.12} & \textbf{35.86} & \textbf{0.9931} \\
\bottomrule
\end{tabular}}
\end{table*}

\begin{table}[ht]
\centering
\caption{Quantitative comparisons on the OpenIllumination dataset. Methods with \textdagger mean the metrics are from the Gaussian Object paper.}
\resizebox{0.5\textwidth}{!}{
\label{tab:2}
\begin{tabular}{l|ccc|ccc}
\hline
\multirow{2}{*}{\textbf{Method}} & \multicolumn{3}{c|}{\textbf{4-view}} & \multicolumn{3}{c}{\textbf{6-view}} \\ 
                & LPIPS$\downarrow$ & PSNR$\uparrow$ & SSIM$\uparrow$ & LPIPS$\downarrow$ & PSNR$\uparrow$ & SSIM$\uparrow$ \\ \hline
DVGO\textdagger            & 11.84           & 21.15           & 0.8973          & 8.83            & 23.79           & 0.9209       \\ 
3DGS\textdagger            & 30.08           & 11.50           & 0.8454          & 29.65           & 11.98           & 0.8277       \\ 
DietNeRF\textdagger & 10.66       & 23.09           & 0.9361         & 9.51            & 24.20           & 0.9401       \\ 
RegNeRF\textdagger & 47.31         & 11.61           & 0.6940          & 30.28           & 14.08           & 0.8586      \\ 
FreeNeRF\textdagger & 35.81         & 12.21           & 0.7969          & 35.15           & 11.47           & 0.8128\\ 
SparseNeRF\textdagger      & 22.28           & 13.60           & 0.8808          & 26.30           & 12.80           & 0.8403      \\ 
ZeroRF\textdagger & 9.74           & 24.54           & 0.9308          & 7.96            & 26.51           & 0.9415      \\ 
G.O.\textdagger        & 6.71            & 24.64           & 0.9354          & 5.44            & 26.54           & 0.9443   \\
\midrule
Ours (w/o PP)        & 5.86            & 24.66           & 0.9372         & 5.08            & 26.55           & 0.9453   \\
Ours (full setting)       & \textbf{5.75}            & \textbf{25.23}           & \textbf{0.9458}         & \textbf{4.79}            & \textbf{27.00}           & \textbf{0.9582}   \\
\hline
\end{tabular}
}
\end{table}

\begin{table}
    \captionof{table}{Ablation study of augmentation strategies conducted on the 'kitchen' dataset with 4 input views.}
    \label{tab:mask}
    \resizebox{0.5\textwidth}{!}{
    \begin{tabular}{@{}lccc@{}}
    \toprule
    Setting & LPIPS* $\downarrow$ & PSNR $\uparrow$ & SSIM $\uparrow$ \\ 
    \midrule
    w/o augmentation & 8.17 & 22.82 & 0.9141 \\
    w/o perceptual view augmentation & 6.76 & 23.98 & 0.9228 \\
    w/o mask augmentation & 6.56 & 24.11 & 0.9229 \\
    Ours (full setting) & \textbf{5.14} & \textbf{26.87} & \textbf{0.9379} \\
    \bottomrule
    \end{tabular}
    }
    \vspace{-5mm}
\end{table}

\begin{figure}
\centering
\begin{tikzpicture}[scale=0.82]
    \begin{axis}[
        ybar=4, 
        enlarge x limits=0.6, 
        legend style={
            at={(0.5,-0.15)},
            anchor=north,
            legend columns=-1
        },
        ylabel={Time (seconds)},
        symbolic x coords={training time, rendering time, total time},
        xtick=data,
        nodes near coords,
        nodes near coords align={vertical},
        bar width=18pt, 
        xtick align=inside, 
        x tick label style={rotate=0,anchor=north} 
    ]
    \addplot coordinates {(training time,7027.8) (rendering time,88.7) };
    \addplot coordinates {(training time,425.1) (rendering time,80.8) };
    \legend{Gaussian Object, AugGaussian}
    \end{axis}
\end{tikzpicture}
\caption{Efficiency comparison under same settings with 4 input views on dataset MipNeRF360.}
\vspace{-4mm}
\label{fig:time-comparison}
\end{figure}
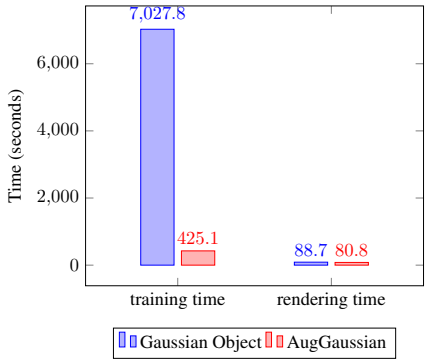

\subsection{Implementation Detail}
Our framework extends 3D Gaussian Splatting (3DGS) \cite{kerbl20233d} with a two-stage approach for sparse-view 3D reconstruction.The first stage involves multiple training iterations, utilizing ZoeDepth \cite{bhat2023zoedepth} for accurate depth estimation. A point-based mask strategy selectively refines critical geometric areas to enhance fidelity.

In the second stage, a fine Gaussian model is trained with a patch-based mask strategy to refine surface details. Using FPS and KNN for point cloud segmentation, it enhances texture and detail. The process takes under 10 minutes on a Tesla T4 GPU with 4, 6, and 9 779×520 images, demonstrating its efficiency for high-fidelity 3D reconstruction from sparse data.

\subsection{Dataset}
We evaluate our method on three datasets for sparse-view 360° object reconstruction: Mip-NeRF360 \cite{Barron2021MipNeRF3U}, OmniObject3D \cite{wu2023omniobject3d}, and OpenIllumination \cite{liu2024openillumination}.
Mip-NeRF360 extends mip-NeRF \cite{DBLP:journals/corr/abs-2103-13415} to handle unbounded scenes using non-linear parameterization and online distillation. OmniObject3D includes over 6,000 real-scanned 3D objects across 190 categories, enhancing 3D representation learning. OpenIllumination, captured with LightStage, is utilized in its sparse form as introduced in ZeroRF \cite{shi2024zerorf}.

\subsection{Evaluation}
We compare our AugGS model to state-of-the-art 3D reconstruction methods, including GaussianObject \cite{yang2024gaussianobject}, 3DGS \cite{kerbl20233d}, DVGO \cite{sun2022direct}, RegNeRF \cite{Niemeyer2021Regnerf}, DietNeRF \cite{jain2021putting}, SparseNeRF \cite{wang2023sparsenerf}, ZeroRF \cite{shi2024zerorf}, and FSGS \cite{zhu2023FSGS}. GaussianObject uses visual hull techniques and a pretrained repair model, while FSGS incorporates SfM-point initialization with additional points for sparse 360° views. Baseline and few-view models employ various regularization strategies. All models are evaluated using their publicly available implementations to ensure consistent comparisons (see Tables \ref{tab:comparisons} and \ref{tab:2}). Our AugGS outperforms GaussianObject on key metrics across three datasets without extra optimizations, demonstrating our lightweight model’s efficiency and superior PSNR for image quality.

Our structural training approach addresses sparse data challenges and scales with more data. On diverse datasets, our model surpasses state-of-the-art benchmarks in PSNR and SSIM. Compared to GaussianObject \cite{yang2024gaussianobject}, we improved LPIPS by 5–30\%, PSNR by 8–25\%, and SSIM by 5–10\%. Qualitative results show smoother, detailed reconstructions of complex structures like gears and flower pots, demonstrating our model’s superior handling of intricate details.

\subsection{Ablation Studies}

In our ablation study, we evaluate four augmentation setups: no augmentation, disabling perceptual view augmentation, disabling mask augmentation, and combining all. As shown in Tab. \ref{tab:mask} and Fig. \ref{fig:mask}, each augmentation improves performance, with the combined approach achieving the highest fidelity. Without all augmentations, the model faile to accurately reconstruct obscured structures like the quarter-exposed gear and vehicle front. Using all strategies enable effective detail capture, highlighting the essential role of augmentation in enhancing structural integrity and surface detail for high-quality 3D reconstructions.

We have also examined how iteration numbers affect model performance on the MipNeRF360 and Omni3D datasets. The coarse Gaussian phase uses groups every 500 iterations, while the fine phase uses every 1000. Optimal coarse iterations range from 1000 to 5000, and fine Gaussian peak around 6000 iterations. More complex structures require additional fine iterations, indicating that structural complexity necessitates tailored iteration counts. Optimal coarse iterations do not guarantee better fine refinement, highlighting the need for a customized approach based on object complexity.

Additionally, we evaluate the efficiency of our AugGS model (Fig. \ref{fig:time-comparison}). Compared to the state-of-the-art GaussianObject \cite{yang2024gaussianobject}, AugGS reduces training time to 6.3\% and overall runtime to 7.1\% by utilizing a simple framework without complex feature extraction or pre-trained models. On the 4-view kitchen dataset, AugGS maintains similar rendering times while requiring less memory and GPU resources. This greater efficiency accelerates the research and development cycle and enables deployment on edge devices, making 3D reconstruction technology more accessible and sustainable for real-world applications.
\section{Conclusion}

This paper presents a sparse-view 3D reconstruction method using self-augmented Gaussian splatting method and structure-aware masks to generate high-fidelity models from limited 2D inputs. Our method improves geometric precision, outperforming state-of-the-art methods with higher efficiency and lower cost.
Future work will optimize modeling, enhance mask generation, and extend to dynamic scenes, setting a new benchmark for sparse-view 3D reconstruction.

\bibliographystyle{IEEEbib}
\bibliography{icme2025references}

\end{document}